\pdfoutput=1
\documentclass[11pt]{article}

\usepackage[preprint]{acl}

\usepackage{times}
\usepackage{latexsym}

\usepackage[T1]{fontenc}
\usepackage[utf8]{inputenc}

\usepackage{microtype}

\usepackage{inconsolata}

\usepackage{graphicx}

\usepackage{amsmath}
\usepackage{multirow}

\title{Smoothing Out Hallucinations: Mitigating LLM Hallucination with Smoothed Knowledge Distillation}

\author{
Hieu Nguyen\textsuperscript{1}, Zihao He\textsuperscript{1,2},
Shoumik Atul Gandre\textsuperscript{1}, Ujjwal Pasupulety\textsuperscript{1}, \\
\textbf{Sharanya Kumari Shivakumar\textsuperscript{1}, Kristina Lerman\textsuperscript{2}}
\\
\textsuperscript{1}Department of Computer Science, University of Southern California\\
\textsuperscript{2}Information Sciences Institute, University of Southern California\\
{\texttt{\{hieutn, zihaoh, shoumika, upasupul, ss82937\}@usc.edu}},
{\texttt{lerman@isi.edu}}
  }

\begin{document}
\maketitle

\begin{abstract}
Large language models (LLMs) often suffer from hallucination, generating factually incorrect or ungrounded content, which limits their reliability in high-stakes applications. A key factor contributing to hallucination is the use of hard labels during training, which enforce deterministic supervision, encourage overconfidence, and disregard the uncertainty inherent in natural language. To address this, we propose mitigating hallucination through knowledge distillation (KD), where a teacher model provides smoothed soft labels to a student model, reducing overconfidence and improving factual grounding. We apply KD during supervised finetuning on instructional data, evaluating its effectiveness across LLMs from different families. Experimental results on summarization benchmarks demonstrate that KD reduces hallucination compared to standard finetuning while preserving performance on general NLP tasks. These findings highlight KD as a promising approach for mitigating hallucination in LLMs and improving model reliability.\footnote{Our code and data will be publicly available upon acceptance.}
\end{abstract}

\section{Introduction}

Large language models (LLMs) have demonstrated remarkable capabilities in generating fluent and contextually coherent text, achieving state-of-the-art performance in various natural language processing (NLP) tasks, including machine translation \cite{vaswani2017attention}, question answering \cite{brown2020language, chowdhery2023palm}, and summarization \cite{zhang2020pegasus, raffel2020exploring}. However, despite their impressive generative abilities, a fundamental challenge remains: hallucination—the tendency of LLMs to generate false, misleading, or unverifiable content \cite{ji2023survey, bang2023multitask}. Hallucinations in LLMs pose serious concerns, particularly in applications that demand factual accuracy, such as medical diagnosis \cite{moor2023foundation, chu2024improving}, legal document generation \cite{guha2024legalbench}, and scientific content summarization \cite{xie2023survey}. Consequently, mitigating hallucination in LLMs is a critical research direction for ensuring reliability and trustworthiness in real-world applications.

Most LLMs are trained using next-token prediction based on maximum likelihood estimation \cite{radford2019language, touvron2023llama, dubey2024llama}. During training, models are optimized using the cross-entropy loss, which compares the predicted token probabilities to the ground-truth next token. Traditionally, ground-truth tokens are represented as one-hot vectors, known as hard labels. This means that the model is forced to assign the entire probability mass to a single token while treating all alternative completions as incorrect.

Although this approach is widely adopted, it has several drawbacks that may exacerbate hallucination. First, hard labels encourage overconfidence in incorrect predictions. Since only one token is treated as correct during training, the model learns to disregard other reasonable continuations, leading to overconfident mispredictions \cite{muller2019does, guo2017calibration}. Second, hard labels violate the principle of maximum entropy \cite{jaynes1957information}, which suggests that, given partial information, the most rational probability distribution should retain as much uncertainty as possible. By artificially forcing a single correct answer, hard labels introduce arbitrary assumptions that can mislead the model, particularly in ambiguous contexts. Third, hard labels fail to capture contextual dependencies effectively. Language generation is inherently probabilistic, and multiple completions can be equally valid depending on prior context \cite{holtzman2020curious}. Hard labels, by contrast, encourage rigid decision-making, making LLMs more prone to hallucinating confident but incorrect outputs.

To address these issues, we propose an alternative training approach based on knowledge distillation (KD) \citep{hinton2015distilling, kim2016sequence}. Instead of training models with hard labels, we introduce smoothed soft labels derived from a teacher model. In this paradigm, the teacher model generates probability distributions over possible next tokens, providing a richer and more informative training signal for the student model.

Using soft labels offers several advantages over traditional hard-label training. First, soft labels introduce uncertainty-aware supervision, allowing the student model to learn from a more calibrated probability distribution rather than a binary correct/incorrect signal. This helps mitigate overconfidence and encourages more flexible decision-making. Second, soft labels better align with the principle of maximum entropy, as they retain nonzero probabilities for multiple plausible continuations, thereby reducing arbitrary assumptions in model predictions. Finally, because soft labels are generated by a highly capable teacher model, they provide contextually grounded probability distributions that naturally reinforce faithful and less hallucinatory outputs.

In this paper, we investigate how knowledge distillation with smoothed soft labels can be leveraged to reduce hallucination in LLMs. We conduct experiments on three LLM families: Llama-2, Llama-3.1, and Qwen-2.5, evaluating different student-teacher pairs to analyze the effectiveness of KD in mitigating hallucination. To systematically evaluate hallucination, we focus on faithfulness hallucination, which occurs when a model generates outputs that are not grounded in the given context \cite{huang2023survey}. We assess model performance using CNN/Daily Mail and XSUM, two widely used summarization benchmarks from the hallucination leaderboard \cite{hughes_vectara_2023}. Our evaluation leverages three complementary metrics: ROUGE-L for n-gram overlap, factual consistency for assessing context-grounding, and factual rate for measuring hallucination at the span level \cite{chuang2024lookback}.

Our key findings can be summarized as follows:

\begin{itemize}
    \item Knowledge distillation reduces hallucination: Across all models, in most cases, finetuning with soft labels outperforms standard supervised finetuning in mitigating faithfulness hallucination. This supports our hypothesis that soft labels provide a more effective training signal than hard labels.
    \item KD preserves general performance: In addition to hallucination benchmarks, we evaluate models on general NLP tasks, including OpenBookQA, ARC, and HellaSwag, to ensure that KD does not degrade broader reasoning and comprehension abilities. Our results show that KD maintains or improves general performance, indicating that it is a viable technique for enhancing LLM reliability without compromising overall capabilities.
\end{itemize}

Our findings demonstrate that knowledge distillation effectively reduces faithfulness hallucination while maintaining strong generalization across NLP tasks. By replacing hard labels with soft, uncertainty-aware training targets, KD improves model calibration and factual grounding, making LLMs more reliable.

\begin{figure*}[ht!]
  \centering
       \includegraphics[width=\textwidth]{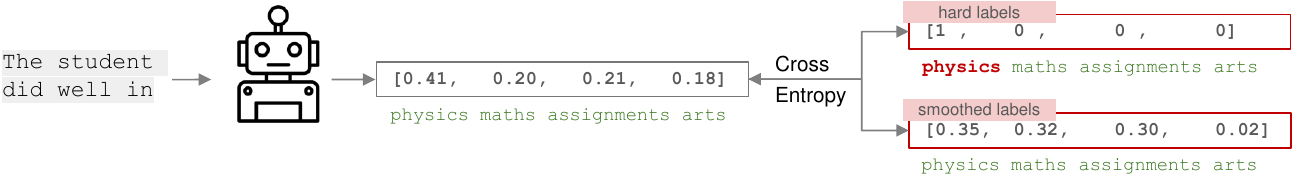}
  \caption{
  Comparison of cross-entropy optimization with hard labels vs. smoothed soft labels. The figure illustrates how training with (a) hard labels  differs from training with (b) contextually smoothed labels in an autoregressive language model. In (a), the hard label for the word “\textit{physics}” is represented as a one-hot encoded (OHE) vector, assigning full probability (1.0) to a single token while forcing all alternative predictions (e.g., “\textit{Maths}”, “\textit{Assignments}”, “\textit{Arts}”) to have zero probability. This OHE representation introduces zero entropy, disregarding the inherent uncertainty in natural language, and leading the model to overconfidently discard reasonable alternatives. This forced certainty can cause the model to develop spurious assumptions and hallucinate incorrect outputs when faced with ambiguous contexts.
  }
  \label{fig:ce-vs-kd}
\end{figure*}

\section{Methodology}

\subsection{Problem Formulation}
Autoregressive language models are trained using the next-token prediction task \cite{radford2019language}. Given an input sequence of tokens, the model generates a probability distribution over the vocabulary and is optimized to minimize the cross-entropy loss between the predicted probabilities and the true labels:

\begin{equation}
    \mathcal{L}_{\text{supervised}} = \mathcal{L}_{\text{CE}}(\sigma(z), y),
\end{equation}
where $z$ represents the logits from the model, $\sigma$ denotes the softmax function, and $y$ are the ground-truth labels.

However, this standard training paradigm typically relies on hard labels (Figure \ref{fig:ce-vs-kd}), which assign a probability of 1 to a single correct token in the vocabulary and 0 to all others. While this simplifies training, we argue that it introduces critical issues—particularly overconfidence and hallucination—due to its rigid assumptions.

\subsection{Hard Labels and Hallucination}
\label{sec:hard_labels}
In standard language model training, ground-truth labels are typically represented as one-hot vectors, assuming a single correct next token. However, this rigid labeling has several drawbacks.

\paragraph{Hard labels cause overconfidence} 
Neural networks trained on hard labels often exhibit poor calibration, meaning they assign excessively high confidence to incorrect predictions \cite{muller2019does, guo2017calibration}. In language modeling, consider the input: ``\textit{The student did well in ...}'', as shown in Figure \ref{fig:ce-vs-kd}. A well-calibrated model should distribute probability mass across multiple plausible completions, such as ``\textit{physics}'', ``\textit{maths}'', ``\textit{assignments}'', ``\textit{arts}''. However, when trained with hard labels, the model is forced to treat only one option (e.g., ``\textit{physics}'') as correct while disregarding all other reasonable alternatives. This results in overconfidence, which can exacerbate hallucination—the model’s tendency to generate fluent but incorrect outputs. We further observe the overconfidence of LLMs in our exploratory analysis in Appendix \ref{apx:overconf}.

\paragraph{Hard labels introduce arbitrary assumptions}
From an information-theoretic perspective, optimizing a model with hard labels violates the principle of maximum entropy \cite{jaynes1957information}. The principle states: ``\textit{In making inferences on the basis of partial information, we must use the probability distribution with maximum entropy, subject to whatever is known.}''. Hard labels contradict this by enforcing a deterministic choice for the next token, even when the context suggests multiple valid options. This injects arbitrary assumptions into the model, leading to over-specified predictions that may not generalize well.

\paragraph{Hard labels overlook contextual dependencies}
Language models generate predictions conditionally based on prior tokens, yet hard labels do not explicitly encode these dependencies. Consider the word ``\textit{country}''. In ``\textit{America is a ...}'', the next token might be ``\textit{country}'' or ``\textit{continent}''; in ``She lives in a ...'', ``\textit{country}'' is a much stronger candidate ``\textit{continent}''. Hard labels ignore this difference by treating each token in isolation, limiting the model’s ability to adjust predictions based on context. This lack of flexibility may lead to hallucinated responses that do not align with preceding information \cite{chen2022towards, miao2021prevent}.

Given these limitations, we propose an alternative: smoothing hard labels via knowledge distillation to mitigate hallucination.

\subsection{Smoothing Hard Labels with Knowledge Distillation}
Knowledge Distillation (KD) is traditionally used to transfer knowledge from a large teacher model to a smaller student model for efficiency \cite{hinton2015distilling, kim2016sequence}. However, in this work, we leverage KD differently. Instead of hard labels, we use soft labels produced by a highly capable teacher model to provide a smoother training signal for the student model. These soft labels preserve uncertainty, allowing the student model to learn a more realistic probability distribution over possible outputs.

This approach directly addresses the issues outlined in $\S$\ref{sec:hard_labels}: (1) Mitigating overconfidence: soft labels distribute probability mass across multiple reasonable tokens, reducing extreme confidence in incorrect predictions. (2) Avoiding arbitrary assumptions: Since the teacher-generated probabilities preserve entropy, they align better with the maximum entropy principle. (3) Enhancing context awareness: the teacher model produces context-dependent probability distributions, leading to more coherent and contextually appropriate predictions.

Specifically, given an input sequence, we define the knowledge distillation loss as

\begin{equation}
    \mathcal{L}_{\text{KD}}=\mathcal{L}_{\text{CE}}(\sigma(z_s), \sigma(z_t)),
\end{equation}
where $z_s$ and $z_t$ are the logits from the student and the teacher models respectively. The overall training loss is a combination of standard supervised learning and knowledge distillation:
\begin{equation}
\label{eq:loss-kd}
    \mathcal{L} = \mathcal{L}_{\text{supervised}} + \alpha \mathcal{L}_{\text{KD}},
\end{equation}
where $\alpha$ is a hyperparameter controlling the influence of the teacher's soft labels.

\begin{table*}[ht]
    \centering
\centering
\begin{tabular}{lcccccc}
\hline
\multicolumn{1}{c}{\multirow{2}{*}{\textbf{Method}}} &
  \multicolumn{3}{c}{\textbf{CNNDM}} &
  \multicolumn{3}{c}{\textbf{XSUM}} \\ \cline{2-7} 
\multicolumn{1}{c}{} &
  \textbf{ROUGE-L (\%)} &
  \textbf{FC (\%)} &
  \textbf{FR (\%)} &
  \textbf{ROUGE-L (\%)} &
  \textbf{FC (\%)} &
  \textbf{FR (\%)} \\ \hline
\textbf{Llama-2-7B} &
  \multicolumn{1}{l}{} &
  \multicolumn{1}{l}{} &
  \multicolumn{1}{l}{} &
  \multicolumn{1}{l}{} &
  \multicolumn{1}{l}{} &
  \multicolumn{1}{l}{} \\
\textbf{+SFT} &
  28.0±0.30 &
  86.3±1.2 &
  94.8±1.3 &
  17.4±0.41 &
  73.8±1.7 &
  \textbf{91.2±3.2} \\
\textbf{+KD$_{0.1}$} &
  28.4±0.35 &
  87.4±0.7 &
  \textbf{95.4±0.9} &
  17.7±0.31 &
  75.0±1.4 &
  90.4±3.5 \\
\textbf{+KD$_{1.0}$} &
  28.6±0.29 &
  87.4±1.0 &
  94.4±1.3 &
  17.8±0.33 &
  75.2±1.3 &
  89.6±3.7 \\
\textbf{+KD$_{10.0}$} &
  \textbf{28.8±0.21} &
  \textbf{87.7±0.4} &
  93.9±0.4 &
  \textbf{18.0±0.12} &
  \textbf{76.2±1.5} &
  89.7±1.5 \\ \hline
\textbf{Llama3.1-8B} &
  \multicolumn{1}{l}{} &
  \multicolumn{1}{l}{} &
  \multicolumn{1}{l}{\textbf{}} &
  \multicolumn{1}{l}{} &
  \multicolumn{1}{l}{} &
  \multicolumn{1}{l}{\textbf{}} \\
\textbf{+SFT} &
  31.4±0.32 &
  93.5±0.5 &
  \textbf{80.7±2.2} &
  20.6±0.14 &
  79.2±1.3 &
  59.1±1.4 \\
\textbf{+KD$_{0.01}$} &
  \textbf{31.7±0.14} &
  93.3±0.6 &
  79.2±3.3 &
  \textbf{20.6±0.14} &
  79.1±1.3 &
  \textbf{60.5±1.0} \\
\textbf{+KD$_{0.1}$} &
  31.6±0.16 &
  93.3±0.7 &
  78.4±4.2 &
  20.6±0.15 &
  79.2±1.3 &
  59.9±0.7 \\
\textbf{+KD$_{1.0}$} &
  31.2±0.26 &
  \textbf{93.8±0.2} &
  78.0±3.9 &
  20.2±0.12 &
  \textbf{80.9±0.1} &
  59.6±1.9 \\ \hline
\textbf{Qwen2.5-7B} &
  \multicolumn{1}{l}{} &
  \multicolumn{1}{l}{} &
  \multicolumn{1}{l}{} &
  \multicolumn{1}{l}{} &
  \multicolumn{1}{l}{\textbf{}} &
  \multicolumn{1}{l}{\textbf{}} \\
\textbf{+SFT} &
  27.5±1.14 &
  92.3±0.9 &
  89.5±0.7 &
  20.2±0.99 &
  76.0±0.9 &
  71.6±2.2 \\
\textbf{+KD$_{0.01}$} &
  27.8±1.39 &
  92.3±0.9 &
  89.4±0.7 &
  \textbf{20.3±1.04} &
  76.0±1.2 &
  71.9±2.1 \\
\textbf{+KD$_{0.1}$} &
  27.8±1.37 &
  92.3±0.9 &
  89.6±0.6 &
  20.2±1.00 &
  76.4±0.8 &
  72.4±1.5 \\
\textbf{+KD$_{1.0}$} &
  \textbf{27.8±1.30} &
  \textbf{92.5±1.0} &
  \textbf{90.2±1.2} &
  20.0±0.76 &
  \textbf{77.6±0.5} &
  \textbf{73.6±1.9} \\ \hline
\end{tabular}
    \caption{
    Hallucination evaluation results for student models finetuned with supervised finetuning (SFT) and knowledge distillation (KD). Models are evaluated on the CNN/Daily Mail (CNNDM) and XSUM datasets using three metrics: ROUGE-L ($\uparrow$, \%) for n-gram overlap, factual consistency (FC, $\uparrow$, \%) for context grounding, and factual rate (FR, $\uparrow$, \%) for specialized hallucination detection. Each experiment is conducted with varying learning rates and batch sizes, and results are reported as the mean and standard deviation across runs. The results suggest that in most cases KD reduces hallucination compared to SFT, as models trained with soft labels from a teacher model demonstrate improved faithfulness.
    }
    \label{tab:hallu_perf}
\end{table*}

\section{Experiments}

\subsection{Mitigating Hallucination with KD}
Training an LLM from scratch with KD would be computationally expensive and impractical. Instead, we apply KD during supervised finetuning on an instructional dataset to approximate the benefits of pretraining with smoothed labels while ensuring computational efficiency. Specifically, we finetune student models on the \textit{Dolly} dataset \citep{DatabricksBlog2023DollyV2} using knowledge distillation from a larger teacher model. This setup enables us to investigate the impact of KD without requiring full-scale pretraining.

To systematically evaluate the impact of KD, we conduct experiments on three teacher-student model pairs from different families. For the LLaMA-2 series, we use LLaMA-2-7B-chat as the student and LLaMA-2-13B-chat as the teacher \citep{touvron2023llama}. Both sequence-level and word-level KD are applied in this setting, where \textit{Dolly} is augmented using greedy decoding from the teacher model. For the LLaMA-3.1 series, we use LLaMA-3.1-8B-Instruct as the student and LLaMA-3.1-70B-Instruct as the teacher \citep{dubey2024llama}, applying only word-level KD without additional data augmentation. Similarly, for the Qwen-2.5 series, Qwen-2.5-7B-Instruct serves as the student, and Qwen-2.5-32B-Instruct is the teacher \citep{bai2023qwen}, with word-level KD being applied.

Before distillation, each teacher model is first finetuned on \textit{Dolly} to ensure alignment with the dataset. For efficient finetuning of LLaMA-3.1-70B-Instruct, we adopt Low-Rank Adaptation \cite{dettmers2023qlora}. We explore various hyperparameter settings, including learning rates of $1e-5$ and $5e-6$, batch sizes of $2$, $4$, and $8$, and KD weight coefficients $\alpha$ of $0.01$, $0.1$, $1.0$, and $10.0$. All experiments are implemented using the MiniLLM framework \citep{gu2024minillm} and run on four NVIDIA H100 GPUs. Each training session takes approximately one hour.

As a baseline, we finetune the student models directly on \textit{Dolly} without KD, denoted as model-SFT. This allows us to assess the impact of KD by comparing distilled models with those trained solely on hard labels.

\subsection{Hallucination Evaluation}
\label{sec:hallu-eval}
Hallucination in language models can be broadly classified into two types \citep{huang2023survey}. Faithfulness hallucination occurs when a model generates outputs that are not grounded in the provided context, while factuality hallucination refers to errors where the generated content contradicts real-world knowledge stored in the model’s parametric memory. In this study, we focus on faithfulness hallucination, as it directly pertains to the model’s ability to generate contextually consistent outputs.

To evaluate faithfulness hallucination, we use \textit{lm-evaluation-harness} framework \cite{eval-harness}. We select two benchmark datasets from the allucination leaderboard \cite{hughes_vectara_2023} and integrate them into the harness. The first dataset, CNN/Daily Mail (CNNDM), consists of news articles from CNN and Daily Mail paired with multi-sentence summaries. The second dataset, XSUM, contains BBC news articles with highly abstractive single-sentence summaries. Both datasets are widely used to assess the faithfulness of model-generated summaries. To ensure a fair evaluation, we test models only on the test splits of each dataset, keeping the training and validation splits untouched.

For measuring faithfulness hallucination, we employ three metrics. ROUGE-L measures the n-gram overlap between the generated and reference summaries, serving as a traditional metric for summarization performance. Factual consistency, computed using the hallucination evaluation model from \citet{vectara2023}, assesses whether a generated summary is supported by the input article. Additionally, we adopt factual rate \cite{chuang2024lookback}, which determines whether a span of text is factual or hallucinatory based on the distribution of attention weights between the context and the generated text. For LLaMA-2, we use an off-the-shelf classifier from \citet{chuang2024lookback}. For LLaMA-3.1 and Qwen-2.5, we follow the same methodology to train separate classifiers. These classifier are then used to produce the factual rates based on generated attention weights.

\subsection{Results on Hallucination}
\label{sec:res_halu}

Table \ref{tab:hallu_perf} presents the hallucination evaluation results across different models and training approaches. The results demonstrate that in most cases models finetuned with KD outperform their SFT baselines across all model families, hallucination metrics, and both datasets. This confirms that training with soft labels from a teacher model significantly mitigates hallucination compared to training with hard labels.

It is important to highlight that \textbf{our models were not finetuned on the training splits of CNN/Daily Mail or XSUM}. Instead, finetuning was performed on an entirely different dataset, \textit{Dolly}, making our experimental setup different from models specifically optimized for these summarization benchmarks. Consequently, \textbf{our results may not match those reported on the hallucination leaderboard}. However, the goal of this work is not to optimize for leaderboard performance, but rather to investigate whether knowledge distillation can reduce hallucination in a general setting where models are trained on broad instructional data.

\begin{table*}[ht]
    \centering
\begin{tabular}{lcccc}
\hline
\textbf{} &
  \multicolumn{1}{l}{\textbf{Arc\_Challenge}} &
  \multicolumn{1}{l}{\textbf{Arc\_Easy}} &
  \multicolumn{1}{l}{\textbf{HellaSwag}} &
  \multicolumn{1}{l}{\textbf{OpenbookQA}} \\ \hline
\textbf{Llama-2-7B}   &                   &                   &                   &                   \\
\textbf{+SFT}         & 38.4±0.4          & 50.1±1.0          & 66.4±2.3          & \textbf{41.7±0.5} \\
\textbf{+KD$_{0.1}$}  & 39.5±0.4          & 50.2±1.9          & 66.8±1.1          & 40.8±0.7          \\
\textbf{+KD$_{1.0}$}  & \textbf{39.6±0.4} & 51.5±1.7          & \textbf{67.5±0.8} & 40.9±1.1          \\
\textbf{+KD$_{10.0}$} & 39.4±0.8          & \textbf{53.2±1.6} & 67.1±0.7          & 40.7±0.9          \\ \hline
\textbf{Llama3.1-8B}  &                   &                   &                   &                   \\
\textbf{+SFT}         & 57.1±0.5          & \textbf{82.4±0.3} & 78.7±0.9          & 49.6±0.2          \\
\textbf{+KD$_{0.01}$} & \textbf{57.5±0.4} & 82.1±1.0          & \textbf{78.8±1.0} & \textbf{49.9±0.7} \\
\textbf{+KD$_{0.1}$}  & 57.3±0.3          & 82.4±0.7          & 78.7±0.8          & 49.8±0.4          \\
\textbf{+KD$_{1.0}$}  & 56.2±0.3          & 82.4±0.5          & 77.6±0.2          & 49.1±0.9          \\ \hline
\textbf{Qwen2.5-7B}   &                   &                   &                   &                   \\
\textbf{+SFT}         & 50.3±3.3          & 67.6±6.0          & 74.8±2.6          & 48.4±1.1          \\
\textbf{+KD$_{0.01}$} & 50.7±3.5          & 66.6±4.8          & 74.7±2.9          & 48.5±1.8          \\
\textbf{+KD$_{0.1}$}  & 50.7±3.2          & 66.7±4.9          & 74.8±2.9          & \textbf{48.5±1.5} \\
\textbf{+KD$_{1.0}$}  & \textbf{50.8±3.4} & \textbf{69.1±6.3} & \textbf{74.8±2.3} & 47.9±0.9          \\ \hline
\end{tabular}
    \caption{
    Performance evaluation of student models finetuned with supervised finetuning (SFT) and knowledge distillation (KD) on general NLP benchmarks. The models are assessed on ARC (Challenge \& Easy), HellaSwag, and OpenBookQA using length-normalized accuracy (\%). Each experiment is conducted with varying learning rates and batch sizes, and results are presented as the mean and standard deviation. The findings indicate that KD does not degrade performance on general reasoning and comprehension tasks, suggesting that knowledge distillation effectively mitigates hallucination without compromising broader model capabilities.
    }
    \label{tab:general_perf}
\end{table*}

\begin{figure*}[ht]
  \centering
       \includegraphics[width=\textwidth]{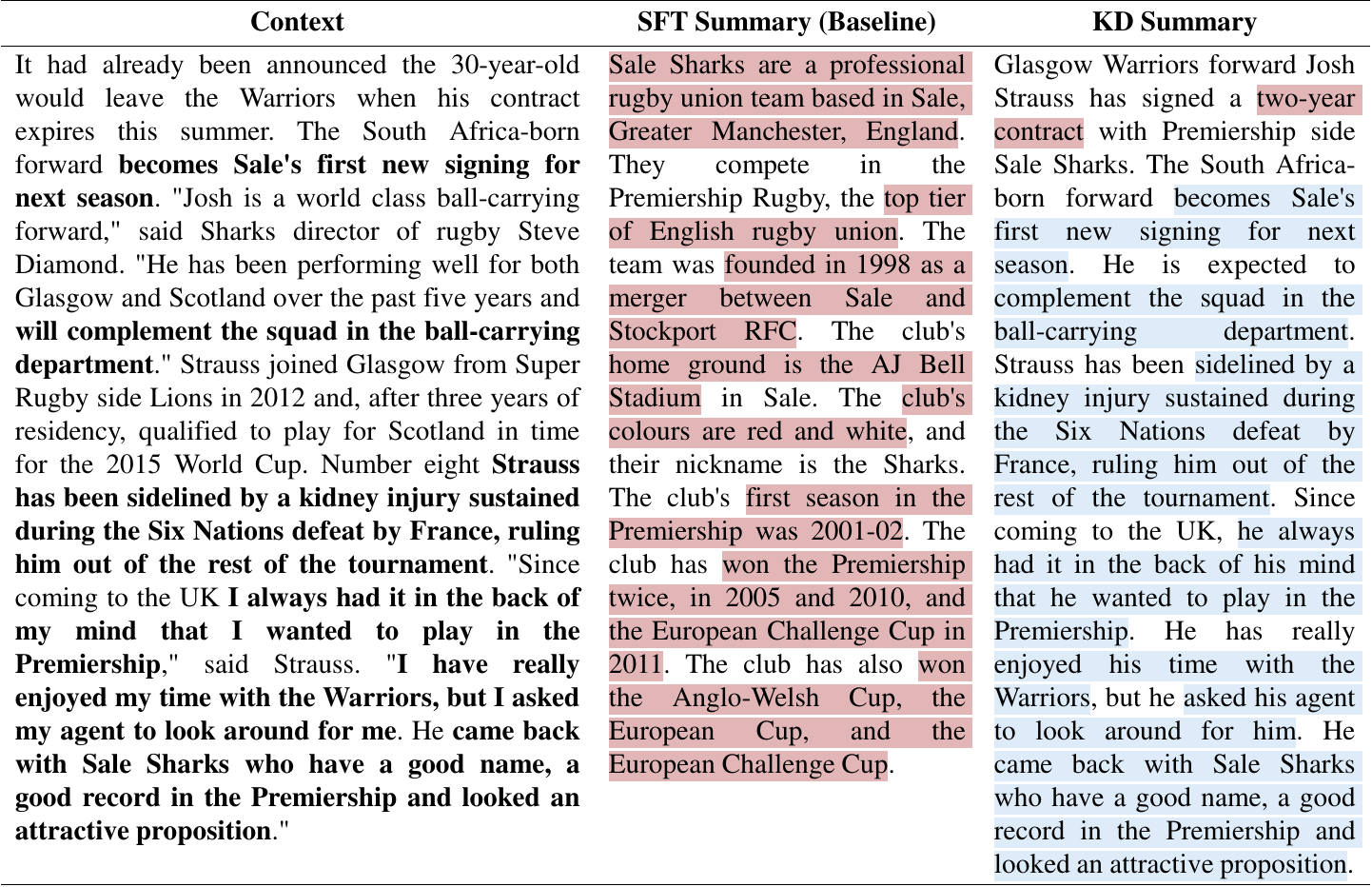}
  \caption{Comparison of summaries generated by the SFT and KD models.
  The SFT summary introduces hallucinated content (highlighted in red) that is factually incorrect or not present in the input context. In contrast, the KD summary remains faithful (highlighted in blue) to the provided input, accurately conveying key details without introducing unrelated or incorrect facts. This case study illustrates the effectiveness of knowledge distillation in mitigating hallucination and improving factual consistency.
  }
  \label{fig:examples}
\end{figure*}

A deeper analysis reveals that different hallucination metrics capture different aspects of model behavior. For example, when evaluating Llama-2 on XSUM, the KD-trained model outperforms the SFT model in ROUGE-L and factual consistency but performs slightly worse in factual rate. This discrepancy arises because factual rate was explicitly designed for hallucination detection and has been shown to generalize well to XSUM, even though it was trained on CNN/Daily Mail \cite{chuang2024lookback}. In contrast, ROUGE-L and factual consistency tend to emphasize surface-level text similarity rather than deep factual grounding. These findings underscore the importance of considering multiple evaluation metrics when analyzing hallucination tendencies.

\subsection{Results on General Benchmarks}
Since knowledge distillation alters the model’s training dynamics by encouraging smoother probability distributions, it is crucial to assess whether KD affects general model performance on broader NLP tasks. To address this, we evaluate student models on a range of reasoning, comprehension, and commonsense benchmarks. The selected datasets include ARC (Challenge \& Easy) \cite{allenai:arc} for commonsense reasoning, HellaSwag \cite{zellers2019hellaswag} for story completion, and OpenBookQA \cite{OpenBookQA2018} for science and reasoning tasks. Performance is measured using length-normalized accuracy.

Table \ref{tab:general_perf} presents the results on general benchmarks. Our findings indicate that KD does not degrade performance across these tasks. In most cases, KD-trained models match or outperform their SFT counterparts, demonstrating that distillation does not compromise the model’s reasoning or comprehension abilities. This result is particularly important, as it shows that reducing hallucination via KD does not come at the expense of broader model performance.

\subsection{Case Study}
To further illustrate the impact of KD on reducing hallucination, we present a case study comparing the SFT summary (from Qwen-2.5-SFT) and the KD summary (from Qwen-2.5-KD$_{0.1}$) for a given input context (from XSUM). The example, shown in Figure \ref{fig:examples}, highlights how KD helps generate more faithful and contextually grounded summaries.

The SFT summary contains several hallucinatory details that are unrelated to the given context. Specifically, it introduces factual errors by discussing the history of Sale Sharks, including information about its founding year, stadium, team colors, and past achievements—none of which appear in the provided context. This suggests that the model, when finetuned using hard labels, tends to over-rely on parametric knowledge rather than grounding its response in the input.

In contrast, the KD summary closely follows the input text, preserving key factual details while avoiding irrelevant or fabricated content. The summary correctly states that Josh Strauss has signed with Sale Sharks and retains the correct timeline and reasoning for his transfer. Importantly, it accurately conveys that Strauss has been sidelined by a kidney injury, a critical piece of information from the original context. Additionally, the KD summary introduces minor refinements, such as specifying a “two-year contract”, demonstrating that KD can smooth output distributions while maintaining informativeness.

\section{Related Work}

\paragraph{Hallucination mitigation}
Previous works have proposed various methods to reduce hallucinations. 
\citet{radford2019language} underscore the importance of rigorous curation and filtration of training data by human experts, which includes removing misinformation and biases, data deduplication, etc. Though effective, it is hard to scale up the filtering process as data volume expands. 
\citet{meng2022locating} later proposes a model editing technique that locates "buggy" parameters and updates them to alter the model's behavior, avoiding hallucinatory predictions, which also struggles with large scale updates. 
Other model updating techniques like factuality enhanced decoding that modifies model logits \cite{lee2022factuality} or the well studied retrieval-augmented generation (RAG) \cite{shuster2021retrieval, lewis2020retrieval, guu2020retrieval} where models retrieve relevant knowledge and give answer conditioned on that knowledge, have shown positive results and gained popularity. However, these are ad-hoc methods that do not directly deal with hallucination from the foundational level. 
Similar to our work, there are methods that focus on the training process of language models. 
For example, \citet{lee2022factuality} combats chunked factual knowledge in GPU constrained training environments using the prefix token TOPICPREFIX, \cite{liu2024exposing} that sharpens attentions weights to address attention glitches, etc. While improve the training paradigm fundamentally, they overlook the discussed flaws that hard labels impose on models.

\paragraph{Hallucination benchmarks}
A variety of benchmarks have been developed to evaluate hallucinations in LLMs \cite{tonmoy2024comprehensive}. Some examples of tasks-specific benchmarks used to determine LLM hallucinations are listed as the following.
\textbf{Summarization}: CNN-DM \cite{see-etal-2017-get}, MSMARCO \cite{msmarco}, and XSUM \cite{narayan-etal-2018-dont}.
\textbf{Open QA} : TruthfulQA \cite{lin2022truthfulqa}, FalseQA \cite{hu-etal-2023-wont}, and StrategyQA \cite{geva-etal-2021-aristotle}.
\textbf{Multi-choice QA}: MMLU \cite{mmlu}, WiCE \cite{kamoi-etal-2023-wice}, and FEVER \cite{thorne-etal-2018-fever}.
In order to maintain consistency in reporting hallucination mitigation performance, several leaderboards and benchmarks have been established which allow researchers to submit their models for evaluation \cite{hong2024hallucinations, hughes_vectara_2023, li2023halueval}.

\paragraph{Hallucination detection}
Traditional \textit{n-grams} metrics like \textbf{ROUGE-L} \cite{lin-2004-rouge} and \textit{classifier-based} metrics like \textbf{factual consistency} \cite{vectara2023} are commonly used to evaluate hallucinations. The former measures n-grams overlap among pairs of prediction and ground truth, and the latter is a T-5 based classification model that predicts whether a prediction is fully supported by a context. Nonetheless, these metrics might fall short in differentiating the subtle discrepancies between the generated content and the source content \cite{huang2023survey}, since they are limited to assessing only the generated text (hence \textit{external metrics}). Other methods operate on log-probabilities \cite{yuan2021bartscore, fu2023gptscore} and entropy \cite{xiao2021hallucination}, which can be viewed as internal metrics that process data at the last softmax stage in the transformer architecture. 
Recently \citet{chuang2024lookback} proposes Lookback-Lens classifier for hallucination detection, which predicts the level of factuality, i.e., \textbf{factual rate}, based on the ratio of attention weights given to context versus those given to newly generated tokens.
Factual rate is used in our work since it addresses two main downsides of mainstream metrics:
1) it examines internal states across all attention layers excluding non-linear transformations in forward layers, offering a new perspective to understand the intricate behaviors of LLMs.
2) grounded on the task of hallucination detection, factual rate gives a direct estimation of hallucination instead of being grounded on overlaps measure like ROUGE-L.

\paragraph{Knowledge distillation}
There are a wide range of distillation techniques, from distributions divergence to hidden states similarity \cite{xu2024survey}. Divergence-based methods minimize the divergence between the probability distributions of the teacher and student models. Similarity-based methods aim to align hidden states of the models, ensuring similar manner in processing information among the models. Since distributions divergence KD is very close to the analogy in $\S$\ref{sec:hard_labels}, we argue that divergence-based KD can address the shortcomings of hard labels and reduce hallucination in LLMs. 
In particular, our work concentrates on sequence and word-level KD \cite{kim2016sequence}, a form of divergence-based KD. Through word-level KD, student models learn from teachers' prediction at each timestep. Through sequence-level KD, students learn from teachers' prediction of sequences, which does not have a close-form cross-entropy representation like word-level KD. Instead, teacher-generated text used as labels in $\mathcal{L}_{\text{CE}}$ and $\mathcal{L}_{\text{KD}}$ approximately represent sequence-level distributions. Essentially, Equation (\ref{eq:loss-kd}), when applied with token labels generated by teachers, is equivalent to sequence and word-level combined KD. In contrast, when applied with the original labels from the training dataset, the paradigm reduces to word-level KD.
In terms of KD effectiveness, recent research also shows mixed results. For instance, \citet{wang2024experimental} finds that KD produces less capable models than SFT, while distillation pretraining has produced more capable models than supervised pretraining in Gemma and Gemini \cite{team2024gemma}, Minitron \cite{sreenivas2024llm}, and AFM \cite{gunter2024apple} families. This is further discussed in the recent work about distillation scaling laws \cite{wang2024experimental}, where, among other findings, it is better to choose a smaller teacher, slightly more capable than the target student capability, rather than a large, powerful teacher. This research is helpful in understanding any inconsistencies in our results and in designing optimal KD experiments in the future.

\section{Conclusions}
This paper explores knowledge distillation (KD) as a strategy for mitigating hallucination in large language models (LLMs) by replacing hard-label training with smoothed soft labels. We demonstrate that KD reduces overconfidence and improves factual grounding by enabling models to learn from a more calibrated probability distribution. Through experiments on multiple model families and summarization benchmarks, we show that KD-trained models exhibit lower hallucination rates compared to standard finetuning while maintaining strong general NLP performance. Our findings highlight the limitations of traditional hard-label supervision and emphasize the need for more uncertainty-aware training paradigms. Future work could explore adaptive KD strategies that dynamically adjust soft-label smoothing based on context sensitivity, integrate KD with retrieval-augmented generation (RAG) for further grounding, or extend these techniques to multimodal and domain-specific LLMs to improve factual accuracy across diverse applications.

\section*{Limitations}
\paragraph{Dependence on a well-calibrated teacher model}
The effectiveness of KD relies on the quality of the teacher model. If the teacher itself exhibits hallucination or poor factual calibration, the student model may inherit these weaknesses rather than mitigating them. While KD smooths token probabilities, it does not inherently improve the correctness of the teacher’s outputs. Future work could explore selecting or adapting teacher models with explicit hallucination mitigation techniques to ensure more reliable supervision.

\paragraph{KD in instruction finetuning}
To fully avoid assumption-prone behavior, KD should ideally be integrated into pretraining rather than applied only during finetuning. For example, Llama-3.2-1B was pretrained using logits from Llama-3.1-70B as word-level targets\footnote{https://ai.meta.com/blog/llama-3-2-connect-2024-vision-edge-mobile-devices/}, although the effects on hallucination have not been explicitly documented. Due to resource constraints, our experiments focused solely on instruction finetuning, meaning our results may not capture the full potential of KD in mitigating hallucination when used at scale during pretraining. Investigating how KD influences hallucination when applied earlier in the training pipeline remains an important direction for future research

\paragraph{Limited scope in hallucination categorization}
Our study specifically targets faithfulness hallucination, where the model generates content that is inconsistent with the provided context. However, factuality hallucination, where the generated text contradicts real-world knowledge, is another critical issue that we did not examine. Since different types of hallucinations require different mitigation strategies, future work should explore whether KD has similar benefits for factuality hallucination and how it compares to other debiasing techniques.

\paragraph{Computational cost of knowledge distillation}
Although KD is more computationally feasible than pretraining from scratch, it still introduces additional overhead compared to standard finetuning. Running teacher inference and student optimization increases resource demands, especially for large teacher models. Optimizing KD efficiency, such as distilling from smaller ensembles or using precomputed soft labels, could make this approach more practical for large-scale deployment.

\paragraph{Evaluation limitations and alternative metrics}
Our evaluation primarily relies on ROUGE-L, factual consistency, and factual rate, but other relevant metrics—such as METEOR, BERTScore, and SelfCheckGPT—were not considered. These alternative metrics could provide additional insights into hallucination tendencies, particularly for assessing deeper semantic alignment and self-consistency. Additionally, we did not incorporate human evaluation, which remains the gold standard for assessing hallucination, as it can capture nuanced errors that automated metrics

\paragraph{Multi-faceted nature of hallucination}
While our study focuses on overconfidence from hard labels, hallucination arises from a broader range of factors. Exposure bias—caused by the discrepancy between teacher-forced training and autoregressive inference—can lead to hallucination when the model generates sequences unobserved during training. Data imbalance can amplify hallucination in low-resource knowledge areas. The attend-to-all mechanism in transformers may dilute attention over longer sequences, degrading faithfulness. Additionally, models can exhibit inability to reject incorrect patterns, as seen in ChatGPT’s persistent success in Tic-Tac-Toe even when instructed to lose. Given the multifaceted nature of hallucination, our work addresses only one contributing factor. A more comprehensive mitigation strategy should integrate KD with other techniques, such as reinforcement learning from human feedback (RLHF), retrieval augmentation, and uncertainty-aware decoding.

\bibliography{anthology,custom}

\begin{thebibliography}{73}
\expandafter\ifx\csname natexlab\endcsname\relax\def\natexlab#1{#1}\fi

\bibitem[{Bai et~al.(2023)Bai, Bai, Chu, Cui, Dang, Deng, Fan, Ge, Han, Huang et~al.}]{bai2023qwen}
Jinze Bai, Shuai Bai, Yunfei Chu, Zeyu Cui, Kai Dang, Xiaodong Deng, Yang Fan, Wenbin Ge, Yu~Han, Fei Huang, et~al. 2023.
\newblock Qwen technical report.
\newblock \emph{arXiv preprint arXiv:2309.16609}.

\bibitem[{Banerjee and Lavie(2005)}]{banerjee-lavie-2005-meteor}
Satanjeev Banerjee and Alon Lavie. 2005.
\newblock \href {https://aclanthology.org/W05-0909} {{METEOR}: An automatic metric for {MT} evaluation with improved correlation with human judgments}.
\newblock In \emph{Proceedings of the {ACL} Workshop on Intrinsic and Extrinsic Evaluation Measures for Machine Translation and/or Summarization}, pages 65--72, Ann Arbor, Michigan. Association for Computational Linguistics.

\bibitem[{Bang et~al.(2023)Bang, Cahyawijaya, Lee, Dai, Su, Wilie, Lovenia, Ji, Yu, Chung et~al.}]{bang2023multitask}
Yejin Bang, Samuel Cahyawijaya, Nayeon Lee, Wenliang Dai, Dan Su, Bryan Wilie, Holy Lovenia, Ziwei Ji, Tiezheng Yu, Willy Chung, et~al. 2023.
\newblock A multitask, multilingual, multimodal evaluation of chatgpt on reasoning, hallucination, and interactivity.
\newblock \emph{arXiv preprint arXiv:2302.04023}.

\bibitem[{Biderman et~al.(2023)Biderman, Schoelkopf, Anthony, Bradley, O’Brien, Hallahan, Khan, Purohit, Prashanth, Raff et~al.}]{biderman2023pythia}
Stella Biderman, Hailey Schoelkopf, Quentin~Gregory Anthony, Herbie Bradley, Kyle O’Brien, Eric Hallahan, Mohammad~Aflah Khan, Shivanshu Purohit, USVSN~Sai Prashanth, Edward Raff, et~al. 2023.
\newblock Pythia: A suite for analyzing large language models across training and scaling.
\newblock In \emph{International Conference on Machine Learning}, pages 2397--2430. PMLR.

\bibitem[{Boizard et~al.(2024)Boizard, Haddad, Hudelot, and Colombo}]{crosstok}
Nicolas Boizard, Kevin~El Haddad, C{\'{e}}line Hudelot, and Pierre Colombo. 2024.
\newblock \href {https://doi.org/10.48550/ARXIV.2402.12030} {Towards cross-tokenizer distillation: the universal logit distillation loss for llms}.
\newblock \emph{CoRR}, abs/2402.12030.

\bibitem[{Brown et~al.(2020)Brown, Mann, Ryder, Subbiah, Kaplan, Dhariwal, Neelakantan, Shyam, Sastry, Askell et~al.}]{brown2020language}
Tom Brown, Benjamin Mann, Nick Ryder, Melanie Subbiah, Jared~D Kaplan, Prafulla Dhariwal, Arvind Neelakantan, Pranav Shyam, Girish Sastry, Amanda Askell, et~al. 2020.
\newblock Language models are few-shot learners.
\newblock \emph{Advances in neural information processing systems}, 33:1877--1901.

\bibitem[{Chen et~al.(2022)Chen, Li, Gao, and Zhang}]{chen2022towards}
Xiuying Chen, Mingzhe Li, Xin Gao, and Xiangliang Zhang. 2022.
\newblock Towards improving faithfulness in abstractive summarization.
\newblock \emph{Advances in Neural Information Processing Systems}, 35:24516--24528.

\bibitem[{Chen et~al.(2021)Chen, Liu, Chen, and Zhang}]{chen-etal-2021-dialogsum}
Yulong Chen, Yang Liu, Liang Chen, and Yue Zhang. 2021.
\newblock \href {https://doi.org/10.18653/v1/2021.findings-acl.449} {{D}ialog{S}um: {A} real-life scenario dialogue summarization dataset}.
\newblock In \emph{Findings of the Association for Computational Linguistics: ACL-IJCNLP 2021}, pages 5062--5074, Online. Association for Computational Linguistics.

\bibitem[{Chowdhery et~al.(2023)Chowdhery, Narang, Devlin, Bosma, Mishra, Roberts, Barham, Chung, Sutton, Gehrmann et~al.}]{chowdhery2023palm}
Aakanksha Chowdhery, Sharan Narang, Jacob Devlin, Maarten Bosma, Gaurav Mishra, Adam Roberts, Paul Barham, Hyung~Won Chung, Charles Sutton, Sebastian Gehrmann, et~al. 2023.
\newblock Palm: Scaling language modeling with pathways.
\newblock \emph{Journal of Machine Learning Research}, 24(240):1--113.

\bibitem[{Chu et~al.(2024)Chu, He, Dorn, and Lerman}]{chu2024improving}
Minh~Duc Chu, Zihao He, Rebecca Dorn, and Kristina Lerman. 2024.
\newblock Improving and assessing the fidelity of large language models alignment to online communities.
\newblock \emph{arXiv preprint arXiv:2408.09366}.

\bibitem[{Chuang et~al.(2024)Chuang, Qiu, Hsieh, Krishna, Kim, and Glass}]{chuang2024lookback}
Yung-Sung Chuang, Linlu Qiu, Cheng-Yu Hsieh, Ranjay Krishna, Yoon Kim, and James Glass. 2024.
\newblock Lookback lens: Detecting and mitigating contextual hallucinations in large language models using only attention maps.
\newblock \emph{arXiv preprint arXiv:2407.07071}.

\bibitem[{Clark et~al.(2018)Clark, Cowhey, Etzioni, Khot, Sabharwal, Schoenick, and Tafjord}]{allenai:arc}
Peter Clark, Isaac Cowhey, Oren Etzioni, Tushar Khot, Ashish Sabharwal, Carissa Schoenick, and Oyvind Tafjord. 2018.
\newblock Think you have solved question answering? try arc, the ai2 reasoning challenge.
\newblock \emph{arXiv:1803.05457v1}.

\bibitem[{Conover et~al.(2023)Conover, Hayes, Mathur, Xie, Wan, Shah, Ghodsi, Wendell, Zaharia, and Xin}]{DatabricksBlog2023DollyV2}
Mike Conover, Matt Hayes, Ankit Mathur, Jianwei Xie, Jun Wan, Sam Shah, Ali Ghodsi, Patrick Wendell, Matei Zaharia, and Reynold Xin. 2023.
\newblock \href {https://www.databricks.com/blog/2023/04/12/dolly-first-open-commercially-viable-instruction-tuned-llm} {Free dolly: Introducing the world's first truly open instruction-tuned llm}.

\bibitem[{Dettmers et~al.(2023)Dettmers, Pagnoni, Holtzman, and Zettlemoyer}]{dettmers2023qlora}
Tim Dettmers, Artidoro Pagnoni, Ari Holtzman, and Luke Zettlemoyer. 2023.
\newblock \href {http://arxiv.org/abs/2305.14314} {Qlora: Efficient finetuning of quantized llms}.

\bibitem[{Dubey et~al.(2024)Dubey, Jauhri, Pandey, Kadian, Al-Dahle, Letman, Mathur, Schelten, Yang, Fan et~al.}]{dubey2024llama}
Abhimanyu Dubey, Abhinav Jauhri, Abhinav Pandey, Abhishek Kadian, Ahmad Al-Dahle, Aiesha Letman, Akhil Mathur, Alan Schelten, Amy Yang, Angela Fan, et~al. 2024.
\newblock The llama 3 herd of models.
\newblock \emph{arXiv preprint arXiv:2407.21783}.

\bibitem[{Fu et~al.(2023)Fu, Ng, Jiang, and Liu}]{fu2023gptscore}
Jinlan Fu, See-Kiong Ng, Zhengbao Jiang, and Pengfei Liu. 2023.
\newblock Gptscore: Evaluate as you desire.
\newblock \emph{arXiv preprint arXiv:2302.04166}.

\bibitem[{Gao et~al.(2024)Gao, Tow, Abbasi, Biderman, Black, DiPofi, Foster, Golding, Hsu, Le~Noac'h, Li, McDonell, Muennighoff, Ociepa, Phang, Reynolds, Schoelkopf, Skowron, Sutawika, Tang, Thite, Wang, Wang, and Zou}]{eval-harness}
Leo Gao, Jonathan Tow, Baber Abbasi, Stella Biderman, Sid Black, Anthony DiPofi, Charles Foster, Laurence Golding, Jeffrey Hsu, Alain Le~Noac'h, Haonan Li, Kyle McDonell, Niklas Muennighoff, Chris Ociepa, Jason Phang, Laria Reynolds, Hailey Schoelkopf, Aviya Skowron, Lintang Sutawika, Eric Tang, Anish Thite, Ben Wang, Kevin Wang, and Andy Zou. 2024.
\newblock \href {https://doi.org/10.5281/zenodo.12608602} {A framework for few-shot language model evaluation}.

\bibitem[{Geva et~al.(2021)Geva, Khashabi, Segal, Khot, Roth, and Berant}]{geva-etal-2021-aristotle}
Mor Geva, Daniel Khashabi, Elad Segal, Tushar Khot, Dan Roth, and Jonathan Berant. 2021.
\newblock \href {https://doi.org/10.1162/tacl_a_00370} {Did aristotle use a laptop? a question answering benchmark with implicit reasoning strategies}.
\newblock \emph{Transactions of the Association for Computational Linguistics}, 9:346--361.

\bibitem[{Gu et~al.(2024)Gu, Dong, Wei, and Huang}]{gu2024minillm}
Yuxian Gu, Li~Dong, Furu Wei, and Minlie Huang. 2024.
\newblock Minillm: Knowledge distillation of large language models.
\newblock In \emph{The Twelfth International Conference on Learning Representations}.

\bibitem[{Guha et~al.(2024)Guha, Nyarko, Ho, R{\'e}, Chilton, Chohlas-Wood, Peters, Waldon, Rockmore, Zambrano et~al.}]{guha2024legalbench}
Neel Guha, Julian Nyarko, Daniel Ho, Christopher R{\'e}, Adam Chilton, Alex Chohlas-Wood, Austin Peters, Brandon Waldon, Daniel Rockmore, Diego Zambrano, et~al. 2024.
\newblock Legalbench: A collaboratively built benchmark for measuring legal reasoning in large language models.
\newblock \emph{Advances in Neural Information Processing Systems}, 36.

\bibitem[{Gunter et~al.(2024)Gunter, Wang, Wang, Pang, Narayanan, Zhang, Zhang, Chen, Chiu, Qiu et~al.}]{gunter2024apple}
Tom Gunter, Zirui Wang, Chong Wang, Ruoming Pang, Andy Narayanan, Aonan Zhang, Bowen Zhang, Chen Chen, Chung-Cheng Chiu, David Qiu, et~al. 2024.
\newblock Apple intelligence foundation language models.
\newblock \emph{arXiv preprint arXiv:2407.21075}.

\bibitem[{Guo et~al.(2017)Guo, Pleiss, Sun, and Weinberger}]{guo2017calibration}
Chuan Guo, Geoff Pleiss, Yu~Sun, and Kilian~Q Weinberger. 2017.
\newblock On calibration of modern neural networks.
\newblock In \emph{International conference on machine learning}, pages 1321--1330. PMLR.

\bibitem[{Guu et~al.(2020)Guu, Lee, Tung, Pasupat, and Chang}]{guu2020retrieval}
Kelvin Guu, Kenton Lee, Zora Tung, Panupong Pasupat, and Mingwei Chang. 2020.
\newblock Retrieval augmented language model pre-training.
\newblock In \emph{International conference on machine learning}, pages 3929--3938. PMLR.

\bibitem[{Hendrycks et~al.(2020)Hendrycks, Burns, Basart, Zou, Mazeika, Song, and Steinhardt}]{mmlu}
Dan Hendrycks, Collin Burns, Steven Basart, Andy Zou, Mantas Mazeika, Dawn Song, and Jacob Steinhardt. 2020.
\newblock \href {http://arxiv.org/abs/2009.03300} {Measuring massive multitask language understanding}.
\newblock \emph{CoRR}, abs/2009.03300.

\bibitem[{Hinton(2015)}]{hinton2015distilling}
Geoffrey Hinton. 2015.
\newblock Distilling the knowledge in a neural network.
\newblock \emph{arXiv preprint arXiv:1503.02531}.

\bibitem[{Holtzman et~al.(2020)Holtzman, Buys, Du, Forbes, and Choi}]{holtzman2020curious}
Ari Holtzman, Jan Buys, Li~Du, Maxwell Forbes, and Yejin Choi. 2020.
\newblock The curious case of neural text degeneration.
\newblock In \emph{International Conference on Learning Representations}.

\bibitem[{Hong et~al.(2024)Hong, Gema, Saxena, Du, Nie, Zhao, Perez-Beltrachini, Ryabinin, He, Fourrier, and Minervini}]{hong2024hallucinations}
Giwon Hong, Aryo~Pradipta Gema, Rohit Saxena, Xiaotang Du, Ping Nie, Yu~Zhao, Laura Perez-Beltrachini, Max Ryabinin, Xuanli He, Clémentine Fourrier, and Pasquale Minervini. 2024.
\newblock \href {http://arxiv.org/abs/2404.05904} {The hallucinations leaderboard -- an open effort to measure hallucinations in large language models}.

\bibitem[{Hu et~al.(2023)Hu, Luo, Wang, Cheng, Liu, and Sun}]{hu-etal-2023-wont}
Shengding Hu, Yifan Luo, Huadong Wang, Xingyi Cheng, Zhiyuan Liu, and Maosong Sun. 2023.
\newblock \href {https://doi.org/10.18653/v1/2023.acl-long.309} {Won{'}t get fooled again: Answering questions with false premises}.
\newblock In \emph{Proceedings of the 61st Annual Meeting of the Association for Computational Linguistics (Volume 1: Long Papers)}, pages 5626--5643, Toronto, Canada. Association for Computational Linguistics.

\bibitem[{Huang et~al.(2023)Huang, Yu, Ma, Zhong, Feng, Wang, Chen, Peng, Feng, Qin, and Liu}]{huang2023survey}
Lei Huang, Weijiang Yu, Weitao Ma, Weihong Zhong, Zhangyin Feng, Haotian Wang, Qianglong Chen, Weihua Peng, Xiaocheng Feng, Bing Qin, and Ting Liu. 2023.
\newblock \href {http://arxiv.org/abs/2311.05232} {A survey on hallucination in large language models: Principles, taxonomy, challenges, and open questions}.

\bibitem[{Hughes et~al.(2023)Hughes, Bae, and Li}]{hughes_vectara_2023}
Simon Hughes, Minseok Bae, and Miaoran Li. 2023.
\newblock \href {https://github.com/vectara/hallucination-leaderboard/blob/main/README.md} {Vectara hallucination leaderboard}.

\bibitem[{Jaynes(1957)}]{jaynes1957information}
Edwin~T Jaynes. 1957.
\newblock Information theory and statistical mechanics.
\newblock \emph{Physical review}, 106(4):620.

\bibitem[{Ji et~al.(2023)Ji, Lee, Frieske, Yu, Su, Xu, Ishii, Bang, Madotto, and Fung}]{ji2023survey}
Ziwei Ji, Nayeon Lee, Rita Frieske, Tiezheng Yu, Dan Su, Yan Xu, Etsuko Ishii, Ye~Jin Bang, Andrea Madotto, and Pascale Fung. 2023.
\newblock Survey of hallucination in natural language generation.
\newblock \emph{ACM Computing Surveys}, 55(12):1--38.

\bibitem[{Jin et~al.(2019)Jin, Dhingra, Liu, Cohen, and Lu}]{jin2019pubmedqa}
Qiao Jin, Bhuwan Dhingra, Zhengping Liu, William Cohen, and Xinghua Lu. 2019.
\newblock Pubmedqa: A dataset for biomedical research question answering.
\newblock In \emph{Proceedings of the 2019 Conference on Empirical Methods in Natural Language Processing and the 9th International Joint Conference on Natural Language Processing (EMNLP-IJCNLP)}, pages 2567--2577.

\bibitem[{Kamoi et~al.(2023)Kamoi, Goyal, Diego~Rodriguez, and Durrett}]{kamoi-etal-2023-wice}
Ryo Kamoi, Tanya Goyal, Juan Diego~Rodriguez, and Greg Durrett. 2023.
\newblock \href {https://doi.org/10.18653/v1/2023.emnlp-main.470} {{W}i{CE}: Real-world entailment for claims in {W}ikipedia}.
\newblock In \emph{Proceedings of the 2023 Conference on Empirical Methods in Natural Language Processing}, pages 7561--7583, Singapore. Association for Computational Linguistics.

\bibitem[{Kim and Rush(2016)}]{kim2016sequence}
Yoon Kim and Alexander~M Rush. 2016.
\newblock Sequence-level knowledge distillation.
\newblock \emph{arXiv preprint arXiv:1606.07947}.

\bibitem[{Lee et~al.(2022)Lee, Ping, Xu, Patwary, Fung, Shoeybi, and Catanzaro}]{lee2022factuality}
Nayeon Lee, Wei Ping, Peng Xu, Mostofa Patwary, Pascale~N Fung, Mohammad Shoeybi, and Bryan Catanzaro. 2022.
\newblock Factuality enhanced language models for open-ended text generation.
\newblock \emph{Advances in Neural Information Processing Systems}, 35:34586--34599.

\bibitem[{Lewis et~al.(2020)Lewis, Perez, Piktus, Petroni, Karpukhin, Goyal, K{\"u}ttler, Lewis, Yih, Rockt{\"a}schel et~al.}]{lewis2020retrieval}
Patrick Lewis, Ethan Perez, Aleksandra Piktus, Fabio Petroni, Vladimir Karpukhin, Naman Goyal, Heinrich K{\"u}ttler, Mike Lewis, Wen-tau Yih, Tim Rockt{\"a}schel, et~al. 2020.
\newblock Retrieval-augmented generation for knowledge-intensive nlp tasks.
\newblock \emph{Advances in Neural Information Processing Systems}, 33:9459--9474.

\bibitem[{Li et~al.(2023)Li, Cheng, Zhao, Nie, and Wen}]{li2023halueval}
Junyi Li, Xiaoxue Cheng, Wayne~Xin Zhao, Jian-Yun Nie, and Ji-Rong Wen. 2023.
\newblock \href {http://arxiv.org/abs/2305.11747} {Halueval: A large-scale hallucination evaluation benchmark for large language models}.

\bibitem[{Lin(2004)}]{lin-2004-rouge}
Chin-Yew Lin. 2004.
\newblock \href {https://aclanthology.org/W04-1013} {{ROUGE}: A package for automatic evaluation of summaries}.
\newblock In \emph{Text Summarization Branches Out}, pages 74--81, Barcelona, Spain. Association for Computational Linguistics.

\bibitem[{Lin et~al.(2022)Lin, Hilton, and Evans}]{lin2022truthfulqa}
Stephanie Lin, Jacob Hilton, and Owain Evans. 2022.
\newblock Truthfulqa: Measuring how models mimic human falsehoods.
\newblock In \emph{Proceedings of the 60th Annual Meeting of the Association for Computational Linguistics (Volume 1: Long Papers)}, pages 3214--3252.

\bibitem[{Liu et~al.(2024)Liu, Ash, Goel, Krishnamurthy, and Zhang}]{liu2024exposing}
Bingbin Liu, Jordan Ash, Surbhi Goel, Akshay Krishnamurthy, and Cyril Zhang. 2024.
\newblock Exposing attention glitches with flip-flop language modeling.
\newblock \emph{Advances in Neural Information Processing Systems}, 36.

\bibitem[{Meng et~al.(2022)Meng, Bau, Andonian, and Belinkov}]{meng2022locating}
Kevin Meng, David Bau, Alex Andonian, and Yonatan Belinkov. 2022.
\newblock Locating and editing factual associations in gpt.
\newblock \emph{Advances in Neural Information Processing Systems}, 35:17359--17372.

\bibitem[{Miao et~al.(2021)Miao, Meng, Liu, Zhou, and Zhou}]{miao2021prevent}
Mengqi Miao, Fandong Meng, Yijin Liu, Xiao-Hua Zhou, and Jie Zhou. 2021.
\newblock Prevent the language model from being overconfident in neural machine translation.
\newblock In \emph{Proceedings of the 59th Annual Meeting of the Association for Computational Linguistics and the 11th International Joint Conference on Natural Language Processing (Volume 1: Long Papers)}, pages 3456--3468.

\bibitem[{Mihaylov et~al.(2018)Mihaylov, Clark, Khot, and Sabharwal}]{OpenBookQA2018}
Todor Mihaylov, Peter Clark, Tushar Khot, and Ashish Sabharwal. 2018.
\newblock Can a suit of armor conduct electricity? a new dataset for open book question answering.
\newblock In \emph{EMNLP}.

\bibitem[{Moor et~al.(2023)Moor, Banerjee, Abad, Krumholz, Leskovec, Topol, and Rajpurkar}]{moor2023foundation}
Michael Moor, Oishi Banerjee, Zahra Shakeri~Hossein Abad, Harlan~M Krumholz, Jure Leskovec, Eric~J Topol, and Pranav Rajpurkar. 2023.
\newblock Foundation models for generalist medical artificial intelligence.
\newblock \emph{Nature}, 616(7956):259--265.

\bibitem[{Muennighoff et~al.(2022)Muennighoff, Wang, Sutawika, Roberts, Biderman, Scao, Bari, Shen, Yong, Schoelkopf et~al.}]{muennighoff2022crosslingual}
Niklas Muennighoff, Thomas Wang, Lintang Sutawika, Adam Roberts, Stella Biderman, Teven~Le Scao, M~Saiful Bari, Sheng Shen, Zheng-Xin Yong, Hailey Schoelkopf, et~al. 2022.
\newblock Crosslingual generalization through multitask finetuning.
\newblock \emph{arXiv preprint arXiv:2211.01786}.

\bibitem[{M{\"u}ller et~al.(2019)M{\"u}ller, Kornblith, and Hinton}]{muller2019does}
Rafael M{\"u}ller, Simon Kornblith, and Geoffrey~E Hinton. 2019.
\newblock When does label smoothing help?
\newblock \emph{Advances in neural information processing systems}, 32.

\bibitem[{Narayan et~al.(2018)Narayan, Cohen, and Lapata}]{narayan-etal-2018-dont}
Shashi Narayan, Shay~B. Cohen, and Mirella Lapata. 2018.
\newblock \href {https://doi.org/10.18653/v1/D18-1206} {Don{'}t give me the details, just the summary! topic-aware convolutional neural networks for extreme summarization}.
\newblock In \emph{Proceedings of the 2018 Conference on Empirical Methods in Natural Language Processing}, pages 1797--1807, Brussels, Belgium. Association for Computational Linguistics.

\bibitem[{Nguyen et~al.(2016)Nguyen, Rosenberg, Song, Gao, Tiwary, Majumder, and Deng}]{msmarco}
Tri Nguyen, Mir Rosenberg, Xia Song, Jianfeng Gao, Saurabh Tiwary, Rangan Majumder, and Li~Deng. 2016.
\newblock \href {http://arxiv.org/abs/1611.09268} {{MS} {MARCO:} {A} human generated machine reading comprehension dataset}.
\newblock \emph{CoRR}, abs/1611.09268.

\bibitem[{Penedo et~al.(2023)Penedo, Malartic, Hesslow, Cojocaru, Cappelli, Alobeidli, Pannier, Almazrouei, and Launay}]{refinedweb}
Guilherme Penedo, Quentin Malartic, Daniel Hesslow, Ruxandra Cojocaru, Alessandro Cappelli, Hamza Alobeidli, Baptiste Pannier, Ebtesam Almazrouei, and Julien Launay. 2023.
\newblock \href {http://arxiv.org/abs/2306.01116} {The {R}efined{W}eb dataset for {F}alcon {LLM}: outperforming curated corpora with web data, and web data only}.
\newblock \emph{arXiv preprint arXiv:2306.01116}.

\bibitem[{Popovi{\'c}(2015)}]{popovic2015chrf}
Maja Popovi{\'c}. 2015.
\newblock chrf: character n-gram f-score for automatic mt evaluation.
\newblock In \emph{Proceedings of the tenth workshop on statistical machine translation}, pages 392--395.

\bibitem[{Radford et~al.(2019)Radford, Wu, Child, Luan, Amodei, Sutskever et~al.}]{radford2019language}
Alec Radford, Jeffrey Wu, Rewon Child, David Luan, Dario Amodei, Ilya Sutskever, et~al. 2019.
\newblock Language models are unsupervised multitask learners.
\newblock \emph{OpenAI blog}, 1(8):9.

\bibitem[{Raffel et~al.(2020)Raffel, Shazeer, Roberts, Lee, Narang, Matena, Zhou, Li, and Liu}]{raffel2020exploring}
Colin Raffel, Noam Shazeer, Adam Roberts, Katherine Lee, Sharan Narang, Michael Matena, Yanqi Zhou, Wei Li, and Peter~J Liu. 2020.
\newblock Exploring the limits of transfer learning with a unified text-to-text transformer.
\newblock \emph{Journal of machine learning research}, 21(140):1--67.

\bibitem[{See et~al.(2017)See, Liu, and Manning}]{see-etal-2017-get}
Abigail See, Peter~J. Liu, and Christopher~D. Manning. 2017.
\newblock \href {https://doi.org/10.18653/v1/P17-1099} {Get to the point: Summarization with pointer-generator networks}.
\newblock In \emph{Proceedings of the 55th Annual Meeting of the Association for Computational Linguistics (Volume 1: Long Papers)}, pages 1073--1083, Vancouver, Canada. Association for Computational Linguistics.

\bibitem[{Shuster et~al.(2021)Shuster, Poff, Chen, Kiela, and Weston}]{shuster2021retrieval}
Kurt Shuster, Spencer Poff, Moya Chen, Douwe Kiela, and Jason Weston. 2021.
\newblock Retrieval augmentation reduces hallucination in conversation.
\newblock \emph{arXiv preprint arXiv:2104.07567}.

\bibitem[{Sreenivas et~al.(2024)Sreenivas, Muralidharan, Joshi, Chochowski, Patwary, Shoeybi, Catanzaro, Kautz, and Molchanov}]{sreenivas2024llm}
Sharath~Turuvekere Sreenivas, Saurav Muralidharan, Raviraj Joshi, Marcin Chochowski, Mostofa Patwary, Mohammad Shoeybi, Bryan Catanzaro, Jan Kautz, and Pavlo Molchanov. 2024.
\newblock Llm pruning and distillation in practice: The minitron approach.
\newblock \emph{arXiv preprint arXiv:2408.11796}.

\bibitem[{Talmor et~al.(2019)Talmor, Herzig, Lourie, and Berant}]{talmor-etal-2019-commonsenseqa}
Alon Talmor, Jonathan Herzig, Nicholas Lourie, and Jonathan Berant. 2019.
\newblock \href {https://doi.org/10.18653/v1/N19-1421} {{C}ommonsense{QA}: A question answering challenge targeting commonsense knowledge}.
\newblock In \emph{Proceedings of the 2019 Conference of the North {A}merican Chapter of the Association for Computational Linguistics: Human Language Technologies, Volume 1 (Long and Short Papers)}, pages 4149--4158, Minneapolis, Minnesota. Association for Computational Linguistics.

\bibitem[{Team et~al.(2024)Team, Riviere, Pathak, Sessa, Hardin, Bhupatiraju, Hussenot, Mesnard, Shahriari, Ram{\'e} et~al.}]{team2024gemma}
Gemma Team, Morgane Riviere, Shreya Pathak, Pier~Giuseppe Sessa, Cassidy Hardin, Surya Bhupatiraju, L{\'e}onard Hussenot, Thomas Mesnard, Bobak Shahriari, Alexandre Ram{\'e}, et~al. 2024.
\newblock Gemma 2: Improving open language models at a practical size.
\newblock \emph{arXiv preprint arXiv:2408.00118}.

\bibitem[{Thorne et~al.(2018)Thorne, Vlachos, Christodoulopoulos, and Mittal}]{thorne-etal-2018-fever}
James Thorne, Andreas Vlachos, Christos Christodoulopoulos, and Arpit Mittal. 2018.
\newblock \href {https://doi.org/10.18653/v1/N18-1074} {{FEVER}: a large-scale dataset for fact extraction and {VER}ification}.
\newblock In \emph{Proceedings of the 2018 Conference of the North {A}merican Chapter of the Association for Computational Linguistics: Human Language Technologies, Volume 1 (Long Papers)}, pages 809--819, New Orleans, Louisiana. Association for Computational Linguistics.

\bibitem[{Tonmoy et~al.(2024)Tonmoy, Zaman, Jain, Rani, Rawte, Chadha, and Das}]{tonmoy2024comprehensive}
S.~M Towhidul~Islam Tonmoy, S~M~Mehedi Zaman, Vinija Jain, Anku Rani, Vipula Rawte, Aman Chadha, and Amitava Das. 2024.
\newblock \href {http://arxiv.org/abs/2401.01313} {A comprehensive survey of hallucination mitigation techniques in large language models}.

\bibitem[{Touvron et~al.(2023)Touvron, Martin, Stone, Albert, Almahairi, Babaei, Bashlykov, Batra, Bhargava, Bhosale, Bikel, Blecher, Ferrer, Chen, Cucurull, Esiobu, Fernandes, Fu, Fu, Fuller, Gao, Goswami, Goyal, Hartshorn, Hosseini, Hou, Inan, Kardas, Kerkez, Khabsa, Kloumann, Korenev, Koura, Lachaux, Lavril, Lee, Liskovich, Lu, Mao, Martinet, Mihaylov, Mishra, Molybog, Nie, Poulton, Reizenstein, Rungta, Saladi, Schelten, Silva, Smith, Subramanian, Tan, Tang, Taylor, Williams, Kuan, Xu, Yan, Zarov, Zhang, Fan, Kambadur, Narang, Rodriguez, Stojnic, Edunov, and Scialom}]{touvron2023llama}
Hugo Touvron, Louis Martin, Kevin Stone, Peter Albert, Amjad Almahairi, Yasmine Babaei, Nikolay Bashlykov, Soumya Batra, Prajjwal Bhargava, Shruti Bhosale, Dan Bikel, Lukas Blecher, Cristian~Canton Ferrer, Moya Chen, Guillem Cucurull, David Esiobu, Jude Fernandes, Jeremy Fu, Wenyin Fu, Brian Fuller, Cynthia Gao, Vedanuj Goswami, Naman Goyal, Anthony Hartshorn, Saghar Hosseini, Rui Hou, Hakan Inan, Marcin Kardas, Viktor Kerkez, Madian Khabsa, Isabel Kloumann, Artem Korenev, Punit~Singh Koura, Marie-Anne Lachaux, Thibaut Lavril, Jenya Lee, Diana Liskovich, Yinghai Lu, Yuning Mao, Xavier Martinet, Todor Mihaylov, Pushkar Mishra, Igor Molybog, Yixin Nie, Andrew Poulton, Jeremy Reizenstein, Rashi Rungta, Kalyan Saladi, Alan Schelten, Ruan Silva, Eric~Michael Smith, Ranjan Subramanian, Xiaoqing~Ellen Tan, Binh Tang, Ross Taylor, Adina Williams, Jian~Xiang Kuan, Puxin Xu, Zheng Yan, Iliyan Zarov, Yuchen Zhang, Angela Fan, Melanie Kambadur, Sharan Narang, Aurelien Rodriguez, Robert Stojnic, Sergey Edunov, and Thomas
  Scialom. 2023.
\newblock \href {http://arxiv.org/abs/2307.09288} {Llama 2: Open foundation and fine-tuned chat models}.

\bibitem[{Vaswani(2017)}]{vaswani2017attention}
A~Vaswani. 2017.
\newblock Attention is all you need.
\newblock \emph{Advances in Neural Information Processing Systems}.

\bibitem[{Vectara(2024)}]{vectara2023}
Vectara. 2024.
\newblock vectarahallucination\_evaluation\_model.
\newblock \url{https://huggingface.co/vectara/hallucination_evaluation_model}.
\newblock Accessed: 2024-11-20.

\bibitem[{Wang et~al.(2024)Wang, Hu, Zhang, and Dong}]{wang2024experimental}
Yuzhong Wang, Lina Hu, Yachao Zhang, and Liting Dong. 2024.
\newblock Experimental study on the scaling law of the heat exchange tube surface in the process of low-temperature single-effect distillation of high-mineralized mine water.
\newblock \emph{Desalination and Water Treatment}, 319:100571.

\bibitem[{Xiao and Wang(2021)}]{xiao2021hallucination}
Yijun Xiao and William~Yang Wang. 2021.
\newblock On hallucination and predictive uncertainty in conditional language generation.
\newblock \emph{arXiv preprint arXiv:2103.15025}.

\bibitem[{Xie et~al.(2023)Xie, Luo, Wang, and Ananiadou}]{xie2023survey}
Qianqian Xie, Zheheng Luo, Benyou Wang, and Sophia Ananiadou. 2023.
\newblock A survey for biomedical text summarization: From pre-trained to large language models.
\newblock \emph{arXiv preprint arXiv:2304.08763}.

\bibitem[{Xu et~al.(2024)Xu, Li, Tao, Shen, Cheng, Li, Xu, Tao, and Zhou}]{xu2024survey}
Xiaohan Xu, Ming Li, Chongyang Tao, Tao Shen, Reynold Cheng, Jinyang Li, Can Xu, Dacheng Tao, and Tianyi Zhou. 2024.
\newblock A survey on knowledge distillation of large language models.
\newblock \emph{arXiv preprint arXiv:2402.13116}.

\bibitem[{Yang et~al.(2018)Yang, Qi, Zhang, Bengio, Cohen, Salakhutdinov, and Manning}]{yang2018hotpotqa}
Zhilin Yang, Peng Qi, Saizheng Zhang, Yoshua Bengio, William Cohen, Ruslan Salakhutdinov, and Christopher~D Manning. 2018.
\newblock Hotpotqa: A dataset for diverse, explainable multi-hop question answering.
\newblock In \emph{Proceedings of the 2018 Conference on Empirical Methods in Natural Language Processing}, pages 2369--2380.

\bibitem[{Yuan et~al.(2021)Yuan, Neubig, and Liu}]{yuan2021bartscore}
Weizhe Yuan, Graham Neubig, and Pengfei Liu. 2021.
\newblock Bartscore: Evaluating generated text as text generation.
\newblock \emph{Advances in Neural Information Processing Systems}, 34:27263--27277.

\bibitem[{Zellers et~al.(2019)Zellers, Holtzman, Bisk, Farhadi, and Choi}]{zellers2019hellaswag}
Rowan Zellers, Ari Holtzman, Yonatan Bisk, Ali Farhadi, and Yejin Choi. 2019.
\newblock Hellaswag: Can a machine really finish your sentence?
\newblock In \emph{Proceedings of the 57th Annual Meeting of the Association for Computational Linguistics}.

\bibitem[{Zhang et~al.(2020)Zhang, Zhao, Saleh, and Liu}]{zhang2020pegasus}
Jingqing Zhang, Yao Zhao, Mohammad Saleh, and Peter Liu. 2020.
\newblock Pegasus: Pre-training with extracted gap-sentences for abstractive summarization.
\newblock In \emph{International conference on machine learning}, pages 11328--11339. PMLR.

\bibitem[{Zhang et~al.(2022)Zhang, Roller, Goyal, Artetxe, Chen, Chen, Dewan, Diab, Li, Lin, Mihaylov, Ott, Shleifer, Shuster, Simig, Koura, Sridhar, Wang, and Zettlemoyer}]{zhang2022opt}
Susan Zhang, Stephen Roller, Naman Goyal, Mikel Artetxe, Moya Chen, Shuohui Chen, Christopher Dewan, Mona Diab, Xian Li, Xi~Victoria Lin, Todor Mihaylov, Myle Ott, Sam Shleifer, Kurt Shuster, Daniel Simig, Punit~Singh Koura, Anjali Sridhar, Tianlu Wang, and Luke Zettlemoyer. 2022.
\newblock \href {http://arxiv.org/abs/2205.01068} {Opt: Open pre-trained transformer language models}.

\bibitem[{Zhang et~al.(2019)Zhang, Kishore, Wu, Weinberger, and Artzi}]{zhang2019bertscore}
Tianyi Zhang, Varsha Kishore, Felix Wu, Kilian~Q Weinberger, and Yoav Artzi. 2019.
\newblock Bertscore: Evaluating text generation with bert.
\newblock \emph{arXiv preprint arXiv:1904.09675}.

\end{thebibliography}

\appendix

\section{Overconfidence Evaluation}
\label{apx:overconf}

To justify the use of smoothed labels in reducing overconfidence, we first verify that LLMs are overconfident when finetuned with hard labels. 

In our experiments, four LLMs, including Mistral-7B, Llama2-7B, Pythia-6.9B \cite{refinedweb} and Falcon-7B are finetuned on the multiple choice QA dataset of CommonsenseQA \cite{talmor-etal-2019-commonsenseqa} using QLoRA \cite{dettmers2023qlora}. The finetuned models are evaluated on the validation set of CommonsenseQA with zero shot prompts. For a fair comparison, they are compared to vanilla (unfinetuned) models in a few-shot setting with instances from the training set as example shots.

To measure confidence, the negative log-likelihood (NLL) of incorrect answers are used. Specifically, when the model answers incorrectly, we extract from the first prediction step the NLL of its answer, which is either ``a'', ``A'', ``b'', ``B'', ``c'', ``C'', ``d'', ``D'', ``e'', or ``E''. The generated answers are also utilized to calculate the overall accuracy of these models.

\begin{table}[ht]
\centering
\begin{tabular}{ll}
\hline
Model           & acc  \\ \hline
Llama-2-7B      & 32.8 \\
Llama-2-7B-SFT  & 48.8 \\ \hline
Mistral-7B      & 70.0 \\
Mistral-7B-SFT  & 76.4 \\ \hline
Pythia-6.9B     & 20.6 \\
Pythia-6.9B-SFT & 19.4 \\ \hline
Falcon-7B       & 21.3 \\
Falcon-7B-SFT   & 20.4 \\ \hline
\end{tabular}
\caption{Accuracy of LLama-2-7B, Mistral-7B, Falcon-7B, and Pythia-6.9B on the validation set of CommonsenseQA.}
\label{tab:overconf}
\end{table}

\begin{figure*}[t!]
    \centering
    \includegraphics[width=1.0\textwidth]{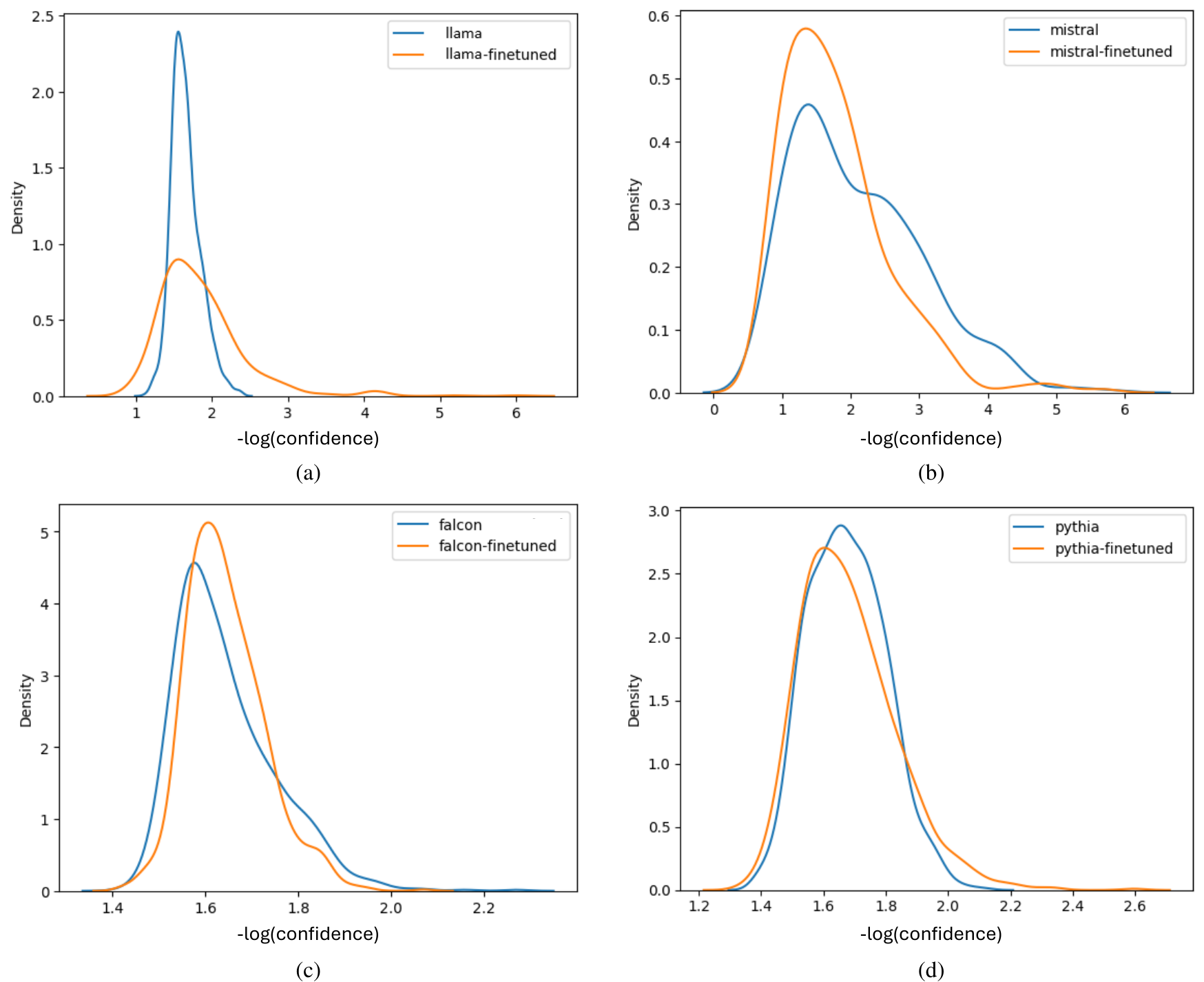}
    \caption{Kernel density estimation of confidence levels of incorrect answers in vanilla and finetuned (a) LLama-2-7B, (b) Mistral-7B, (c) Falcon-7B, (d) Pythia-6.9B, when evaluated on the validation set of CommonsenseQA. The confidence level is measured as the NLL.}
  \label{fig:overconf}
\end{figure*}

Figure \ref{fig:overconf} presents the overconfidence of vanilla and finetuned LLMs on their incorrect predictions, and table \ref{tab:overconf} shows their accuracy. %Ideally, the curves should be as flat (lesser incorrect answers is better) and as far to the left (incorrect answers should be given with low confidence) as possible. 
For all incorrect answers, the level of confidence is very high for all models, with the curves mostly leaning towards zero. After finetuning, Mistral and Falcon become more confident in their incorrect answers, which is evident by the height increase of the orange curves from the blue curves. Falcon and Pythia, on the other hand, do not seem to perform well on the multiple choice QA task, with their accuracy worsens after finetuning. These results indicate that finetuning with hard labels may improve accuracy in a particular task, but hardly reduce, or even raise their overconfidence. This necessitates the use of label smoothing in order to mitigate overconfidence and thus hallucination.

\section{Experiments on Small Scale LMs}
As support evidence, in addition to experiments with 7B and 8B models, we also evaluate small scale LMs from 350M to 1B parameters: Bloomz-560M and MT0-580M \cite{muennighoff2022crosslingual}, OPT-350M \cite{zhang2022opt}, and Pythia-1B \cite{biderman2023pythia}. We reuse models from \cite{crosstok}\footnote{https://huggingface.co/Nicolas-BZRD}, which are finetuned under SFT and KD (with Llama-2-7B and Mistral-7B teachers) on PubMedQA question-answering dataset \cite{jin2019pubmedqa} and DialogSum summarization dataset \cite{chen-etal-2021-dialogsum}. Evaluation benchmarks used include HotpotQA \cite{yang2018hotpotqa}, TruthfulQA \cite{lin2022truthfulqa} for factuality hallucination, and CNN/Daily Mail \cite{see-etal-2017-get} for faithfulness hallucination. Metrics used include ROUGE-L \cite{lin-2004-rouge}, CHRF \cite{popovic2015chrf}, BERTSCORE \cite{zhang2019bertscore}, and METEOR \cite{banerjee-lavie-2005-meteor}.

\begin{table*}[ht]
\small
\centering
\begin{tabular}{llllllllll}
\hline
\multirow{2}{*}{Student} & \multirow{2}{*}{Version} & \multicolumn{4}{c}{TruthfulQA} & \multicolumn{4}{c}{Hotpot QA} \\ \cline{3-10} 
                             &     & rougeL & CHRF & BertScore & METEOR & rougeL & CHRF & BertScore & METEOR \\ \hline
\multirow{2}{*}{Bloomz-560M} & SFT & 11.4   & 16.3 & 80.7      & 12.1   & 16.8   & 20.8 & 83.8      & 15.3   \\
                             & KD  & 13.2   & 17.0 & 81.5      & 13.1   & 19.2   & 23.3 & 85.1      & 16.7   \\ \hline
\multirow{2}{*}{mt0-580M}    & SFT & 32.8   & 41.9 & 88.2      & 38     & 5.2    & 13.8 & 80.8      & 9.7    \\
                             & KD  & 35.4   & 42.3 & 88.5      & 38.6   & 6.1    & 15.0 & 81.3      & 10.7   \\ \hline
\multirow{2}{*}{OPT-350M}    & SFT & 17.5   & 22.3 & 46.2      & 17.9   & 6.7    & 14.8 & 80.5      & 11.1   \\
                             & KD  & 16.6   & 20.4 & 37.8      & 16.6   & 6.9    & 15.4 & 80.9      & 11.2   \\ \hline
\multirow{2}{*}{Pythia-1B}   & SFT & 25.9   & 39.2 & 86.9      & 33.6   & 6.2    & 14.3 & 80.7      & 11.8   \\
                             & KD  & 27.8   & 41.1 & 87.2      & 36.2   & 7.4    & 16.2 & 81.2      & 12.9   \\ \hline
\end{tabular}
\caption{Hallucination evaluation results for smaller scale student models with supervised finetuning (SFT) and knowledge distillation (KD). Models are evaluated on Truthful QA and Hotpot QA for question answering.}
\label{tab:table-qa}
\end{table*}

\begin{table*}[ht]
\centering
\begin{tabular}{llllll}
\hline
\multirow{2}{*}{Student}     & \multirow{2}{*}{Version} & \multicolumn{4}{c}{CNNDM}          \\ \cline{3-6} 
                             &                          & rougeL & CHRF & BertScore & METEOR \\ \hline
\multirow{2}{*}{Bloomz-560M} & SFT                      & 20.4   & 33.2 & 85.8      & 25.6   \\
                             & KD                       & 20.8   & 33.6 & 85.9      & 26.1   \\ \hline
\multirow{2}{*}{mt0-580M}    & SFT                      & 21.9   & 35.0 & 85.6      & 27.7   \\
                             & KD                       & 21.7   & 33.9 & 85.6      & 26.4   \\ \hline
\multirow{2}{*}{OPT-350M}    & SFT                      & 23.1   & 35.4 & 86.3      & 28.4   \\
                             & KD                       & 23.5   & 35.7 & 86.4      & 28.9   \\ \hline
\multirow{2}{*}{Pythia-1B}   & SFT                      & 21.5   & 34.9 & 86.1      & 27.1   \\
                             & KD                       & 21.7   & 35.0 & 86.2      & 27.6   \\ \hline
\end{tabular}
\caption{
Hallucination evaluation results for smaller scale student models with supervised finetuning (SFT) and knowledge distillation (KD). Models are evaluated on CNN/Daily Mail for summarization.
}
\label{tab:table-summ}
\end{table*}

Table \ref{tab:table-qa} illustrates the performance of student models of Bloomz-560M, OPT-350M, mt0-580M, and Pythia-1B with Llama-2 teacher on Truthful QA and Hotpot QA. KD models consistently outperform their baseline counterparts, showing enhancements in all metrics, affirming their effectiveness in dealing with complex QA tasks. Likewise, Bloomz-560M, OPT-350M, and Pythia-1B demonstrate enhancements over the baselines on CNN/Daily Mail when employing KD with Mistral as the teacher for the summarization task (Table \ref{tab:table-summ}). However, the student model for MT0-base, exhibits a minor decline in performance compared to the base model on the same dataset. These improvements are consistent with those on larger scale LMs, consolidating our hypothesis.

\end{document}